\documentclass[runningheads]{llncs}
\usepackage[T1]{fontenc}
\usepackage{graphicx}

\begin{document}
\title{A Mobile Food Recognition System for \\ Dietary Assessment}
\author{Şeymanur Aktı\inst{1}%\orcidID{0000-1111-2222-3333} 
\and
Marwa Qaraqe\inst{2}%\orcidID{1111-2222-3333-4444}
\and
Hazım Kemal Ekenel\inst{1}%\orcidID{2222--3333-4444-5555}
}
\authorrunning{Ş. Aktı et al.}
\institute{Istanbul Technical University, Turkey \\
\email{\{akti15,ekenel\}@itu.edu.tr}\and
Hamad bin Khalifa University, Qatar \\
\email{mqaraqe@hbku.edu.qa}
}
\maketitle              % typeset the header of the contribution
\begin{abstract}

Food recognition is an important task for a variety of applications, including managing health conditions and assisting visually impaired people. Several food recognition studies have focused on generic types of food or specific cuisines, however, food recognition with respect to Middle Eastern cuisines has remained unexplored. %Thus, we improved the study of ~\cite{qaraqe2020automatic} on Middle Eastern cuisine food recognition task and conducted several experiments on the proposed dataset.
Therefore, in this paper we focus on developing a mobile friendly, Middle Eastern cuisine focused food recognition application for assisted living purposes. In order to enable a low-latency, high-accuracy food classification system, we opted to utilize the Mobilenet-v2 deep learning model. As some of the foods are more popular than the others, the number of samples per class in the used Middle Eastern food dataset is relatively imbalanced. To compensate for this problem, data augmentation methods are applied on the underrepresented classes. Experimental results show that using Mobilenet-v2 architecture for this task is beneficial in terms of both accuracy and the memory usage. With the model achieving 94\% accuracy on 23 food classes, the developed mobile application has potential to serve the visually impaired in automatic food recognition via images.

%in the dataset to avoid the performance drop caused by class imbalance. During the experimentation, we proved that these two pre-processing steps have improved the performance of the model. We showed that  The experimental results also show that the Mobilenet-v2 architecture performs as good as, or even better than ResNet-50 and VGG-16 which are more complex architectures. Based on this efficient deep learning model, we also developed a mobile food recognition application. The processed dataset and the experimental results are expected to set a baseline for food recognition on Middle Eastern cuisine task. 

\end{abstract}
\section{Introduction}
The World Health Organization (WHO) estimates that at least 2.2 billion people are classified to be near or distance vision impaired~\cite{who_2021}. The blind and visually impaired face challenges that are often overlooked to sighted people. Research in assistive technology has gained tremendous attention in the past decades to enable people that are visually impaired or blind to perform tasks comparable to sighted people, such as using a phone or computer (via screen reader). However, there still exists many challenges to visually impaired and blind. Among them is the simple task of identifying objects in their presence, which often requires visually impaired people to invoke other senses, such as touch, smell, and taste, to identify such objects. 

With the abundance of food in today’s society as well as the availability of international food locally, it becomes necessary to develop intelligent technology that can aid in the identification of food. Such systems are known as food recognition systems and can identify the type of food using images. Food recognition can be embedded in mobile systems for a diverse set of applications ranging from food tracking for diet management to an assistive technology application to aid visually impaired users identify foods via an image. Such a system will enable people with visual impairment to independently identify various food in real-time, which becomes particularly important i) if they are traveling to foreign countries and would like to identify various meals they will consume, ii) scan menus with images to identify foods in various food delivery applications or in menus at restaurants, and finally iii) to assist in food tracking for various reasons and objectives. 

There has been work done in the literature that focuses on the development of food recognition systems through images. However, the majority of the work conducted on food recognition focuses on western style foods or popular international foods, such as sushi, ramen, etc. Middle Eastern cuisine has always been underrepresented in such food recognition applications, and limited work has investigated within this scope. Among the first to propose a food recognition system catering to foods communally consumed in the Middle East is~\cite{qaraqe2020automatic}. In this paper, the authors implement a machine learning approach to develop the food recognition algorithm on a novel dataset collected by the authors. Their model is based on a feature extraction and classification stage and explore the capabilities of both early and late fusion techniques. In the early fusion technique, the authors combine the extracted features using different combination techniques. Ultimately, an accuracy of 80\% was achieved on the developed dataset. 
%We developed a mobile application

%Summary of the work and contributions
In this work, we also focus on Middle Eastern cuisine, and aim at enhancing the results of the food recognition model presented in~\cite{qaraqe2020automatic}. The proposed approach led to a 10\% absolute increase in the accuracy while also being more computationally efficient compared to the previous work. %Based on this efficient deep learning based method, a mobile food recognition application will be developed and tested.

%Our contributions are as follows:

%\let\labelitemi\labelitemii

%\begin{itemize}
 %   \item Processing the Middle Eastern food dataset~\cite{qaraqe2020automatic} by combining the classes with high similarity and applying data augmentation to reduce the effect of the class imbalance.
  %  \item Experimenting with Mobilenet-v2~\cite{sandler2018mobilenetv2} on the processed dataset and performance assessment.
   % \item Developing a mobile application utilizing the deep learning model fine-tuned on the food dataset.
%\end{itemize}

%The second aim of this work is to leverage the power of deep learning in image classification tasks. In particular, MobileNet-v2~\cite{sandler2018mobilenetv2} will be used for the image-based classification  as it has a relatively simple architecture, allowing for real-time image inference, while maintaining   promising performance.

\section{Related Work}

As nutrition is one of the most important factors for human development and health, it has become a subject for computer vision and deep learning fields. Researchers have been interested in food-related problems, such as estimating taste appreciation~\cite{zulfikar2016preliminary}, generating pizza recipes~\cite{papadopoulos2019make}, comprehension of cooking recipes~\cite{yagcioglu2018recipeqa}, recipe generation from food images~\cite{salvador2019inverse}, and food portion estimation~\cite{fang2018single}. There has also been many systems developed for food image classification utilizing mobile devices~\cite{zhao2020jdnet}, wearable sensors~\cite{liu2012intelligent}, and egocentric cameras~\cite{jia2019automatic}.

There has been some work done for recognition of food images using computer vision methods and hand-crafted features such as SIFT (Scale-invariant Feature Transform)~\cite{lowe2004distinctive} features and color histograms~\cite{chen2009pfid}, bag-of-words~\cite{zong2010combination,anthimopoulos2014food},  pair-wise feature distribution of local features~\cite{yang2010food}, bag-of-features with SIFT, color histogram and Gabor texture feature descriptors~\cite{joutou2009food}.

Given the advancement in intelligent algorithms, there has been a shift from the manual extraction of features from images to the automatic learning of such features by deep neural networks. As such, deep learning methods have proven their success on image classification tasks in many domains, including food recognition. In specific,  \cite{hassannejad2016food} fine-tuned Inception-v3 architecture~\cite{szegedy2016rethinking} on food images. The authors in \cite{ciocca2017learning} employed a deep residual network~\cite{he2016deep} for training large food datasets, whereas \cite{pandey2017foodnet} and \cite{arslan2021fine} used ensembled deep learning models for food recognition. Instead of using the deep convolutional networks for both extracting features and classifying them,~\cite{mcallister2018combining} used convolutional networks only for feature extraction and used supervised machine learning methods for the classification of the features. A novel network developed specifically for food recognition is proposed in \cite{martinel2018wide} where they exploit the vertical food traits which is common in the available food datasets. Similarly,~\cite{mezgec2017nutrinet} introduced the NutriNet as a novel deep convolutional network architecture for food recognition from images. Considering that the new food images could be included in the learning instead of using static datasets, \cite{He_2021_ICCV} proposed an online continual learning method for food recognition. In order to develop more suitable food recognition models for mobile environments, \cite{zhao2020jdnet} and \cite{heng2021compact} employed knowledge distillation methods resulting in more compact models without losing the performance. The work in \cite{aguilar2021uncertainty} introduced a new data augmentation methodology for food datasets and exploited active learning for food classification to have a deeper insight regarding useful samples in the data. Finally, \cite{nagarajan2021s2ml} tackled the multi-label food classification problem using a novel transfer learning framework.

The available datasets which are also used by some of the works above mostly focus on more generic type of foods~\cite{min2020isia,kaur2019foodx} or specific cuisines such as fast-food image dataset~\cite{chen2009pfid}, Japanese cuisine~\cite{joutou2009food}, Chinese cuisine~\cite{chen2017chinesefoodnet}, western cuisine~\cite{bossard2014food} and more. Different from these datasets, in our study, we worked on Middle Eastern cuisine and improved the work presented in~\cite{qaraqe2020automatic}.

\begin{figure}
\centering
\includegraphics[width=0.9\textwidth]{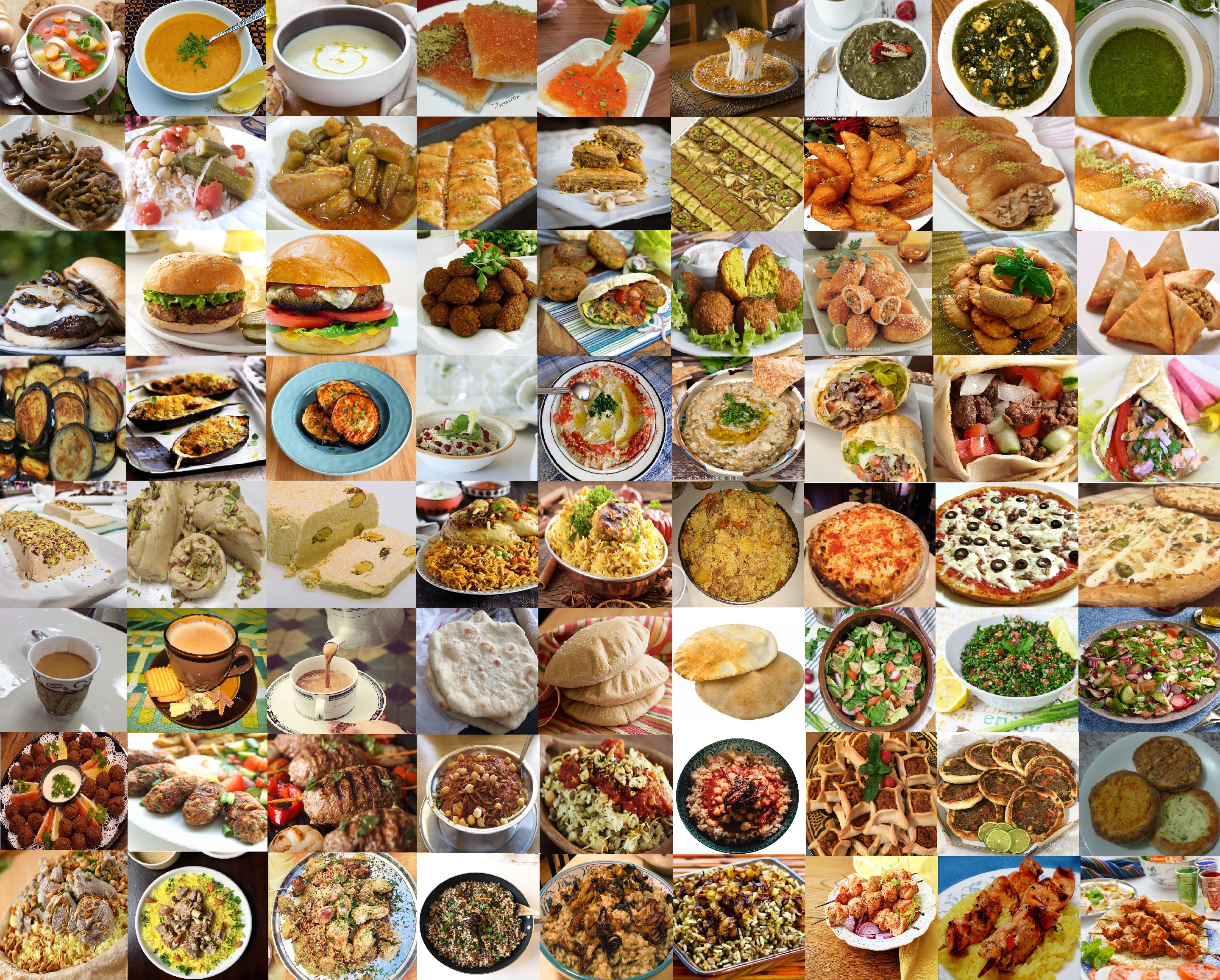}
\caption{Sample images from the Middle Eastern cuisine dataset}
\label{dataset_samples}
\end{figure}

\section{Method}

For the identification of food items via images in real-time, an image classifier, thus, a deep learning model is trained for the classification of 23 classes of food images representing diverse Middle Eastern foods. The employed deep learning model architecture is chosen as a small and portable model providing a low process time for real-time inference. This section discusses the dataset and pre-processing measures taken, as well as the model development for the automatic food recognition task.

\subsection{Dataset and Analysis}
The dataset used was collected by \cite{qaraqe2020automatic} through crawling images from Google and Instagram that represent real-world Middle Eastern cuisine dishes with 27 classes~\cite{qaraqe2020automatic}.  Sample images from the dataset are shown in Figure~\ref{dataset_samples}.
 After examining the dataset, it was found that this dataset has two main challenges, namely, i)  low inter-class difference between similar dishes and ii) class imbalance. To account for these limitations, two pre-processing steps are applied on the dataset prior to training, as discussed below.

\subsubsection{Class Consolidation} 
\begin{figure}
\centering
\includegraphics[width=0.7\textwidth]{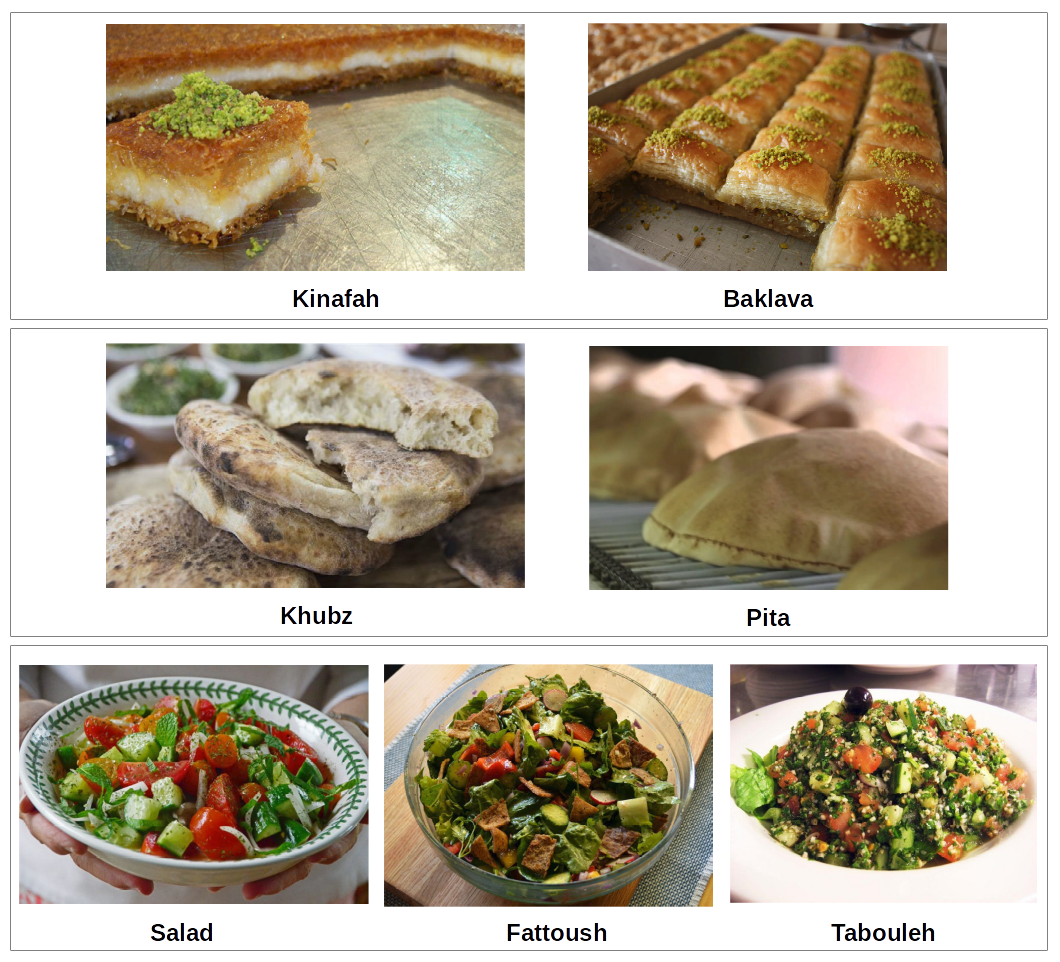}
\caption{Samples from the combined classes}
\label{combined}
\end{figure}

Visual similarity between some of the classes in the food dataset was really high, and many of these classes have similar calorie values and ingredients used. %Given that the initial trained model on whole set of classes had relatively more misclassifications for the similar classes. 
Therefore, a subset of these classes are combined and assigned to a single class. The consolidated classes are listed below and examples from the combined classes can be seen in Figure~\ref{combined}.

\let\labelitemi\labelitemii

\begin{itemize}
    %\item Soup: Arabic soup and hasaa classes are combined since they both refer to the same dish.
     \item Baklava \& Kinafah: Baklava and kinafah are similar kind of dishes which are made of pistachio and syrup with similar calorie content. There exists a high visual similarity between these two classes as the color scales are the same most of the time. Thus, these two classes are combined in order to decrease the number of mispredictions among them.
    \item Khubz \& Pita: These two classes are forms of bread. The bread type consumed widely in Middle Eastern countries is named as pita; however there are small variations of pita bread, such as khubz. As such, these classes were also consolidated.
    %\item Pizza: Types of pizza is a too fine-grained problem for the context of this application and causes many mispredictions. Thus, margherita, chicken pizza and fish pizza classes are combined into a single class. 
    %\item Shawarma: Similar to the pizza class, two types of shawarma, beef shawarma and chicken shawarma are combined into a single class as it is hard to distinguish between these two. 
    \item Salad: There are different types of salads in Middle Eastern cuisine such as tabouleh and fattoush. These salad types differ according to the way they are cut and the dressing, but their ingredients are fairly similar and they have similar calorie content and have high visual similarities. Thus, salad, tabouleh and fattoush classes are combined into a single class. 
    
\end{itemize}

\subsubsection{Data Augmentation} 
\begin{figure}
\includegraphics[width=\textwidth]{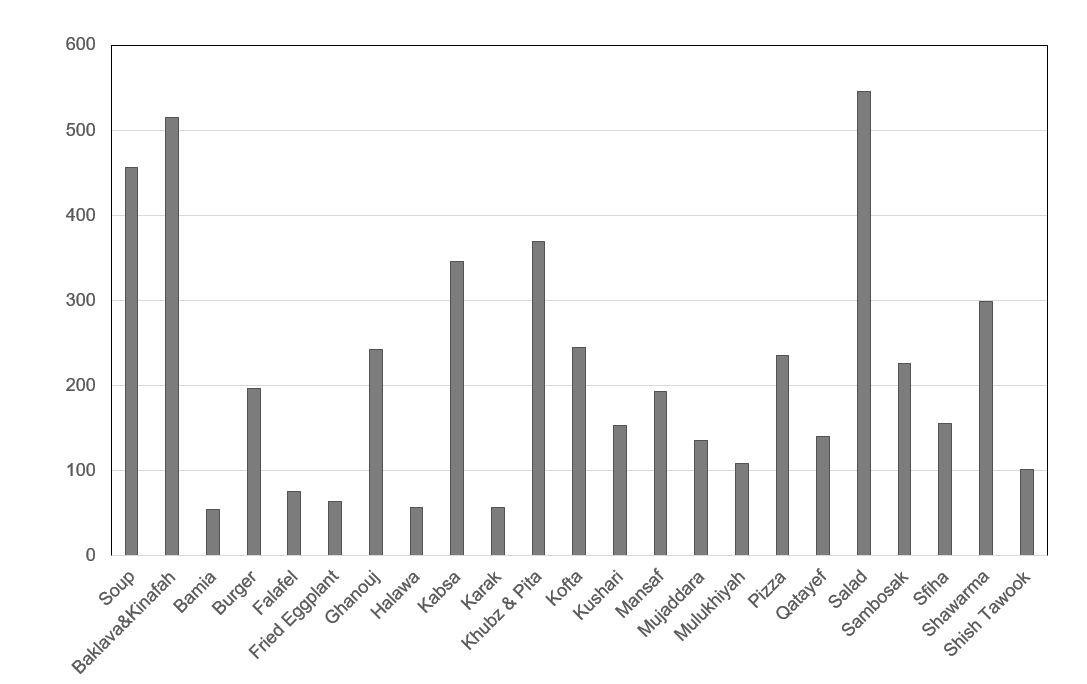}
\caption{Number of samples in each class before applying data augmentation.}
\label{dataset_classes}
\end{figure}

Another limitation with the dataset at hand was the imbalanced number of samples between classes. The distribution of the number of samples per class is given in Figure~\ref{dataset_classes}. In order to overcome the class imbalance problem, some data augmentation techniques are applied on the underrepresented classes to reach a balanced dataset. Data augmentation is only applied on the classes which have less than 100 samples. The applied augmentation methods include horizontal flip, random crop, Gaussian noise, affine transforms, and contrast change, and are designed to generate images from different angles and in different lighting conditions, simulating real images that can be taken using a mobile phone.

%\let\labelitemi\labelitemii

%\begin{itemize}
    %\item Horizontal flip: Flipping the images horizontally simulates different shooting angles.
    %\item Random crop: Applying random crop simulates the variation of shooting distances.
    %\item Gaussian noise: Adding noise simulates the variation of the image quality and blur.
    %\item Affine transforms: Applying affine transforms, we intend to include the augmented images that have some distortions which simulates the images taken from different angles.
    %\item Contrast change: Contrast change simulates the variation in lighting conditions and image quality.
%\end{itemize}

\begin{figure}
\centering
\includegraphics[width=\textwidth]{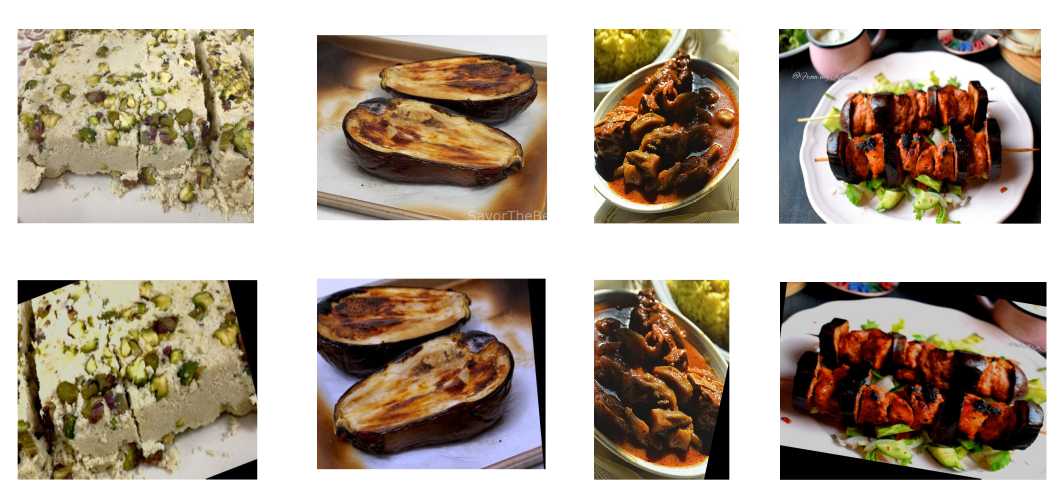}
\caption{Augmented image samples. Original images are at the top and corresponding augmentations are given at the bottom}
\label{augmentations}
\end{figure}

These augmentation steps are applied in sequence to the images where their parameters are selected randomly for each run. Thus, various augmentations are obtained as it can be observed in Figure~\ref{augmentations}. The dataset sizes before and after applying the data augmentation methods are presented in Table~\ref{dataset_table}.

\begin{table}
\centering
\caption{Dataset size before and after data augmentation.}
\begin{tabular}{lccc}
\hline
& Train Set & Test Set & Total \\
\hline
Without augmentation &  4368 & 489 & 4857 \\
With augmentation &  5234 & 489 & 5723\\
\hline
\end{tabular}
\label{dataset_table}
\end{table}

\subsection{Learning Model}

The deep learning architecture for image classification model is chosen as Mobilenet-v2~\cite{sandler2018mobilenetv2}. This model is found to be suitable for our problem where the main aim is twofold, i) predicting the type of the food from the food images with high accuracy, and ii) the model should be lightweight and allow for low-latency predictions on a smartphone. Mobilenet-v2 is a relatively shallow architecture and is more convenient to adapt into the applications that might be used on mobile devices. Nevertheless, it maintains a promising performance for the image classification task, which makes it a good choice for a high-accuracy, low-latency food recognition solution.

A pre-trained model on ImageNet~\cite{deng2009imagenet}
is fine-tuned on the food recognition dataset. During fine-tuning, the last fully-connected layer of the model is replaced with a new fully-connected layer with the output size of 23, as the total number of classes in the dataset is 23. There are 20 blocks in the Mobilenet-v2 architecture and during training the weights for the first ten blocks are frozen and only the last ten blocks are fine-tuned on the food dataset. %to make the model less complex and prevent the model to overfit on the training set.
The learning rate for the last layer's parameters is 1e-03 and the learning rate for other layers' parameters is set to 1e-04. The cross-entropy loss and Adam optimizer are used with batch size of 128. The model is implemented with PyTorch and trained on Nvidia TitanX.

\section{Experiments and Results}

\subsection{Experimental Setup}
For the experimental analysis, the dataset is split into training and test sets. The training set includes 90\% of the original samples. The remaining 10\% is used in the test set. After augmentation, there were a total of 5234 samples in the training set and 489 samples in the test set, as given in Table~\ref{dataset_table}. For the validation, 10-fold cross-validation is used. 

\subsection{Results}

The classification results of the deep learning models are presented in Table~\ref{accuracy-results} in terms of top-1 and top-5 accuracies. The Mobilenet-v2 model trained on the pre-processed dataset gives a satisfying top-1 accuracy where 94\% of the test images are predicted correctly. Additionally, the model achieves a nearly perfect top-5 accuracy, showing that the correct label is found within the high confidence predictions almost all the time. %This outcome is beneficial for the mobile application in terms of the user experience. %In order to avoid the inconveniences, in case of a misprediction, we planned to offer the user to choose the class of the food image among the top-5 predictions. Having a high top-5 accuracy as 99.5\%, this feature is expected to prevent the mistakes caused by mispredictions and improve the application.

In order to assess the benefit of using Mobilenet-v2 for this task, we compared its performance with other larger networks such as ResNet-50~\cite{he2016deep} and VGG-16~\cite{simonyan2014very}. The results are listed in Table~\ref{accuracy-results}. Mobilenet-v2 performs comparable to ResNet-50 and significantly outperforms VGG-16. Moreover, the fine-tuned ResNet-50 model is ten times larger than the Mobilenet-v2 model in terms of the size, which makes the Mobilenet-v2 more efficient considering both the performance and compatibility with mobile devices. As we mentioned before, the food classification module is planned to be employed in the mobile devices, thus using a more compact model is an essential factor for an effective solution. 

\begin{table}
\centering
\caption{Results of the food image classification models}
\begin{tabular}{lcc}
\hline
& Top-1 accuracy & Top-5 accuracy\\
\hline

VGG-16       & 86.7\% & 97.5\% \\
ResNet-50    & 94.0\% & 99.3\% \\
\textbf{Mobilenet-v2} & \textbf{94.0\%} & \textbf{99.5\%} \\

\hline
\end{tabular}
\label{accuracy-results}
\end{table}

 In order to evaluate the effect of the adopted data pre-processing on the performance, several experiments have been conducted. The results of these experiments are presented in Table~\ref{data-process-results}, where there are three studied scenarios; namely, (1) the Mobilenet-v2 architecture is trained without combining the classes that have high visual similarities, (2) Mobilenet-v2 architecture is trained using only the original data, without any data augmentation, (3) Mobilenet-v2 architecture trained on the pre-processed dataset as proposed in Section 3.1. The comparison between scenario (1) and scenario (3) indicates that the low inter-class difference between the food image classes drastically affects the performance as the top-1 accuracy increases by 4\% as we combine the classes having high visual similarities. This is expected since most of the mispredicted validation samples in the initial training of the model on the original data were belonging to these classes. 
 
 The effect of data augmentation, although not significant, is still seen.  Since the data augmentation is applied only the classes that have less than 100 samples in it, the amount of the augmented samples are relatively small and this can explain the limited contribution. Still, applying data augmentation compensates the negative effect of the class imbalance problem to some degree.

%Class accuracies 

\begin{table}
\centering
\caption{Results of the experiments before and after the data pre-processing.}\label{tab1}
\begin{tabular}{lc}
\hline
& Top-1 accuracy \\
\hline
Mobilenet-v2 (w/o combined classes, w/ augmentation) & 90.0\% \\
Mobilenet-v2 (w/ combined classes, w/o augmentation) & 93.5\% \\
Mobilenet-v2 (w/ augmentation and combined classes) & 94.0\% \\

\hline
\end{tabular}
\label{data-process-results}
\end{table}

\begin{figure}
\centering
\includegraphics[width=0.9\textwidth]{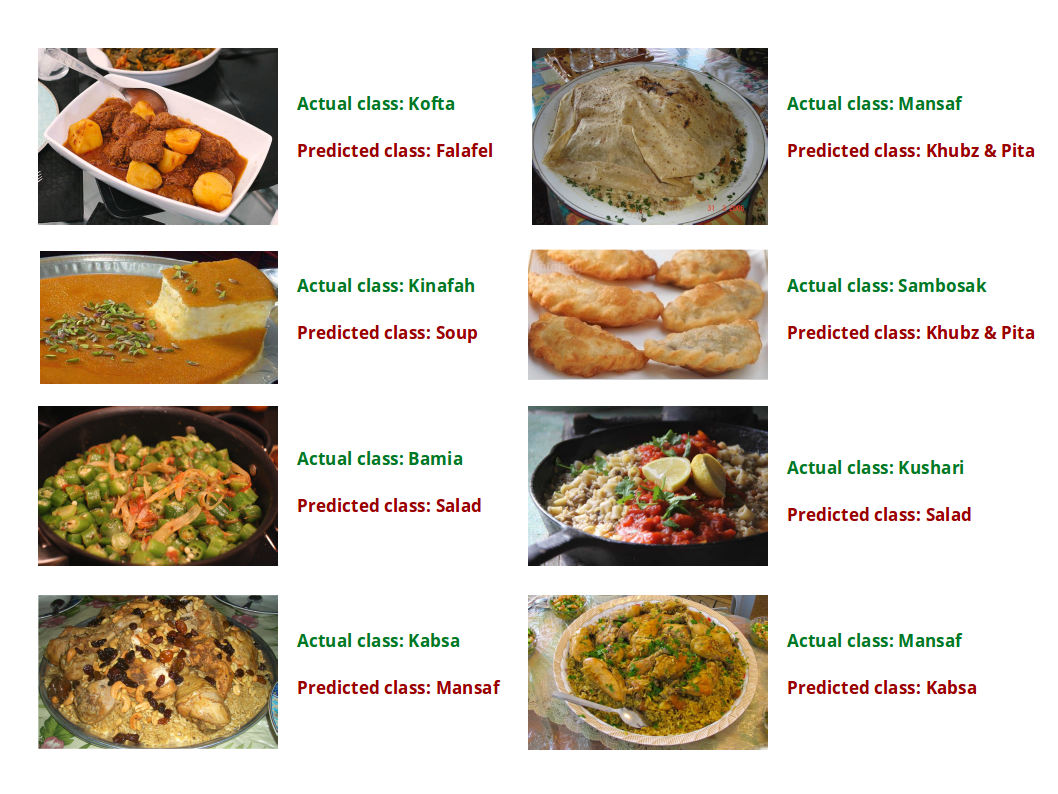}
\caption{Sample mispredictions}
\label{mispredictions}
\end{figure}

As a qualitative analysis, some misclassified samples are given in Figure~\ref{mispredictions}. It can be observed that the mispredictions are mostly reasonable where the misclassified class samples and the ground truth class samples have similar visual patterns, such as kabsa and mansa or kofta and falafel classes. In addition, some of the images may contain overlapping dishes, such as the mansaf with khubz on top of it, thus causing misclassifications.  

\section{Conclusion}

Food classification from images is seen as an important task for health and dietary assessment applications and assisting visually impaired people. The previous studies on food recognition mostly focus on generic types of food or specific cuisines whereas the application for Middle Eastern food recognition remained unexplored. Thus, in this paper, we have improved upon the study of ~\cite{qaraqe2020automatic} on Middle Eastern cuisine food recognition task and conducted several experiments on the proposed dataset. Due to the high inter-class similarity between certain classes, we consolidated some classes that have similar characteristics. Also, some data augmentation methods were applied on the underrepresented classes in the dataset to avoid the performance drop caused by class imbalance. During the experimentation, we proved that these two pre-processing steps have improved the performance of the model. We showed that using Mobilenet-v2 architecture for this task is beneficial in terms of both accuracy and the memory usage. The experimental results also show that the Mobilenet-v2 architecture performs as good as, or even better than ResNet-50 and VGG-16 which are more complex architectures.  The processed dataset and the experimental results are expected to set a baseline for food recognition on Middle Eastern cuisine task. %and lead more enhanced studies on food recognition on mobile devices.

%\subsubsection{Acknowledgements} Please place your acknowledgments at the end of the paper, preceded by an unnumbered run-in heading (i.e. 3rd-level heading).

\bibliographystyle{splncs04}
\bibliography{samplepaper}

\begin{thebibliography}{10}
\providecommand{\url}[1]{\texttt{#1}}
\providecommand{\urlprefix}{URL }
\providecommand{\doi}[1]{https://doi.org/#1}

\bibitem{who_2021}
Vision impairment and blindness (Oct 2021),
  \url{https://www.who.int/news-room/fact-sheets/detail/blindness-and-visual-impairment}

\bibitem{aguilar2021uncertainty}
Aguilar, E., Nagarajan, B., Khantun, R., Bola{\~n}os, M., Radeva, P.:
  Uncertainty-aware data augmentation for food recognition. In: Intl.
  Conference on Pattern Recognition (ICPR) (2021)

\bibitem{anthimopoulos2014food}
Anthimopoulos, M.M., Gianola, L., Scarnato, L., Diem, P., Mougiakakou, S.G.: A
  food recognition system for diabetic patients based on an optimized
  bag-of-features model. IEEE Journal of Biomedical and Health Informatics
  \textbf{18}(4),  1261--1271 (2014)

\bibitem{arslan2021fine}
Arslan, B., Memis, S., Battinisonmez, E., Batur, O.Z.: Fine-grained food
  classification methods on the {UEC Food-100} database. IEEE Transactions on
  Artificial Intelligence  (2021)

\bibitem{bossard2014food}
Bossard, L., Guillaumin, M., Gool, L.V.: Food-101--mining discriminative
  components with random forests. In: European Conference on Computer Vision
  (2014)

\bibitem{chen2009pfid}
Chen, M., Dhingra, K., Wu, W., Yang, L., Sukthankar, R., Yang, J.: {PFID}:
  {P}ittsburgh fast-food image dataset. In: Intl. Conference on Image
  Processing (ICIP) (2009)

\bibitem{chen2017chinesefoodnet}
Chen, X., Zhu, Y., Zhou, H., Diao, L., Wang, D.: Chinesefoodnet: A large-scale
  image dataset for chinese food recognition. arXiv preprint arXiv:1705.02743
  (2017)

\bibitem{ciocca2017learning}
Ciocca, G., Napoletano, P., Schettini, R.: Learning {CNN}-based features for
  retrieval of food images. In: Intl. Conference on Image Analysis and
  Processing (2017)

\bibitem{deng2009imagenet}
Deng, J., Dong, W., Socher, R., Li, L.J., Li, K., Fei-Fei, L.: Imagenet: {A}
  large-scale hierarchical image database. In: Computer Vision and Pattern
  Recognition (2009)

\bibitem{fang2018single}
Fang, S., Shao, Z., Mao, R., Fu, C., Delp, E.J., Zhu, F., Kerr, D.A., Boushey,
  C.J.: Single-view food portion estimation: {L}earning image-to-energy
  mappings using generative adversarial networks. In: Intl. Conference on Image
  Processing (ICIP) (2018)

\bibitem{hassannejad2016food}
Hassannejad, H., Matrella, G., Ciampolini, P., De~Munari, I., Mordonini, M.,
  Cagnoni, S.: Food image recognition using very deep convolutional networks.
  In: Intl. Workshop on Multimedia Assisted Dietary Management (2016)

\bibitem{He_2021_ICCV}
He, J., Zhu, F.: Online continual learning for visual food classification. In:
  Intl. Conference on Computer Vision (ICCV) Workshops (2021)

\bibitem{he2016deep}
He, K., Zhang, X., Ren, S., Sun, J.: Deep residual learning for image
  recognition. In: Computer Vision and Pattern Recognition (2016)

\bibitem{heng2021compact}
Heng, Z., Yap, K.H., Kot, A.C.: A compact joint distillation network for visual
  food recognition. In: Intl. Conference on Acoustics, Speech and Signal
  Processing (ICASSP) (2021)

\bibitem{jia2019automatic}
Jia, W., Li, Y., et~al.: Automatic food detection in egocentric images using
  artificial intelligence technology. Public Health Nutrition  \textbf{22}(7),
  1168--1179 (2019)

\bibitem{joutou2009food}
Joutou, T., Yanai, K.: A food image recognition system with multiple kernel
  learning. In: Intl. Conference on Image Processing (ICIP) (2009)

\bibitem{kaur2019foodx}
Kaur, P., Sikka, K., Wang, W., Belongie, S., Divakaran, A.: Foodx-251: a
  dataset for fine-grained food classification. arXiv preprint arXiv:1907.06167
   (2019)

\bibitem{liu2012intelligent}
Liu, J., Johns, E., Atallah, L., Pettitt, C., Lo, B., Frost, G., Yang, G.Z.: An
  intelligent food-intake monitoring system using wearable sensors. In: Intl.
  Conference on Wearable and Implantable Body Sensor Networks (2012)

\bibitem{lowe2004distinctive}
Lowe, D.G.: Distinctive image features from scale-invariant keypoints. Intl.
  Journal of Computer Vision  \textbf{60}(2),  91--110 (2004)

\bibitem{martinel2018wide}
Martinel, N., Foresti, G.L., Micheloni, C.: Wide-slice residual networks for
  food recognition. In: Winter Conference on Applications of Computer Vision
  (WACV) (2018)

\bibitem{mcallister2018combining}
McAllister, P., Zheng, H., Bond, R., Moorhead, A.: Combining deep residual
  neural network features with supervised machine learning algorithms to
  classify diverse food image datasets. Computers in Biology and Medicine
  \textbf{95},  217--233 (2018)

\bibitem{mezgec2017nutrinet}
Mezgec, S., Korou{\v{s}}i{\'c}~Seljak, B.: Nutri{N}et: {A} deep learning food
  and drink image recognition system for dietary assessment. Nutrients
  \textbf{9}(7), ~657 (2017)

\bibitem{min2020isia}
Min, W., Liu, L., Wang, Z., Luo, Z., Wei, X., Wei, X., Jiang, S.: Isia
  food-500: A dataset for large-scale food recognition via stacked global-local
  attention network. In: ACM Intl. Conference on Multimedia (2020)

\bibitem{nagarajan2021s2ml}
Nagarajan, B., Aguilar, E., Radeva, P.: {S2ML-TL} framework for multi-label
  food recognition. In: Intl. Conference on Pattern Recognition (2021)

\bibitem{pandey2017foodnet}
Pandey, P., Deepthi, A., Mandal, B., Puhan, N.B.: Food{N}et: {R}ecognizing
  foods using ensemble of deep networks. IEEE Signal Processing Letters
  \textbf{24}(12),  1758--1762 (2017)

\bibitem{papadopoulos2019make}
Papadopoulos, D.P., Tamaazousti, Y., Ofli, F., Weber, I., Torralba, A.: How to
  make a pizza: Learning a compositional layer-based gan model. In: Computer
  Vision and Pattern Recognition (2019)

\bibitem{qaraqe2020automatic}
Qaraqe, M., Usman, M., Ahmad, K., Sohail, A., Boyaci, A.: Automatic food
  recognition system for middle-eastern cuisines. IET Image Processing
  \textbf{14}(11),  2469--2479 (2020)

\bibitem{salvador2019inverse}
Salvador, A., Drozdzal, M., Gir{\'o}-i Nieto, X., Romero, A.: Inverse cooking:
  {R}ecipe generation from food images. In: Computer Vision and Pattern
  Recognition (2019)

\bibitem{sandler2018mobilenetv2}
Sandler, M., Howard, A., Zhu, M., Zhmoginov, A., Chen, L.C.: Mobilenetv2:
  {I}nverted residuals and linear bottlenecks. In: Computer Vision and Pattern
  Recognition (2018)

\bibitem{simonyan2014very}
Simonyan, K., Zisserman, A.: Very deep convolutional networks for large-scale
  image recognition. arXiv preprint arXiv:1409.1556  (2014)

\bibitem{szegedy2016rethinking}
Szegedy, C., Vanhoucke, V., Ioffe, S., Shlens, J., Wojna, Z.: Rethinking the
  inception architecture for computer vision. In: Computer Vision and Pattern
  Recognition (2016)

\bibitem{yagcioglu2018recipeqa}
Yagcioglu, S., Erdem, A., Erdem, E., Ikizler-Cinbis, N.: Recipe{QA}: {A}
  challenge dataset for multimodal comprehension of cooking recipes. In:
  Empirical Methods in Natural Language Processing (2018)

\bibitem{yang2010food}
Yang, S., Chen, M., Pomerleau, D., Sukthankar, R.: Food recognition using
  statistics of pairwise local features. In: Computer Society Conference on
  Computer Vision and Pattern Recognition (2010)

\bibitem{zhao2020jdnet}
Zhao, H., Yap, K.H., Kot, A.C., Duan, L.: Jdnet: {A} joint-learning distilled
  network for mobile visual food recognition. IEEE Journal of Selected Topics
  in Signal Processing  \textbf{14}(4),  665--675 (2020)

\bibitem{zong2010combination}
Zong, Z., Nguyen, D.T., Ogunbona, P., Li, W.: On the combination of local
  texture and global structure for food classification. In: Intl. Symposium on
  Multimedia. pp. 204--211. IEEE (2010)

\bibitem{zulfikar2016preliminary}
Z{\"u}lfikar, {\.I}.E., Dibeklio{\u{g}}lu, H., Ekenel, H.K.: A preliminary
  study on visual estimation of taste appreciation. In: Intl. Conference on
  Multimedia \& Expo Workshops (ICMEW) (2016)

\end{thebibliography}

\end{document}